\documentclass[10pt,conference]{IEEEtran}

\usepackage[utf8]{inputenc} 
\usepackage{multirow}
\usepackage{booktabs}
\usepackage{graphicx}
\usepackage{array}
\usepackage{tcolorbox}
\usepackage{color, colortbl}
\usepackage{amsmath}
\usepackage{xspace}
\usepackage{enumitem}
\usepackage{hyperref}
\usepackage{listings}
\usepackage{pifont}
\usepackage{soul}
\usepackage[dvipsnames]{xcolor}
\usepackage{url}
\usepackage{subfigure}
\usepackage[ruled,vlined]{algorithm2e}
\usepackage{tabularx}
\usepackage{colortbl}
\usepackage{threeparttable}
\usepackage{caption}
\usepackage{array}
\usepackage{fontawesome5}
\usepackage{balance}
\usepackage{float}

\newtcolorbox{finding}[1]{
  colframe=green!40!black, 
  colback=green!5!white,
  coltitle=white,
  fonttitle=\bfseries,
  title={\faLightbulb\hspace{0.5em}\textbf{#1}},
  sharp corners,
  boxrule=1pt,
  boxsep=1mm, 
  left=2mm,   
  right=2mm,  
  top=1mm,    
  bottom=1mm  
}

\definecolor{headergray}{gray}{0.95}
\definecolor{codegreen}{rgb}{0,0.6,0}      
\definecolor{codegray}{rgb}{0.5,0.5,0.5}       
\definecolor{codepurple}{rgb}{0.58,0,0.82}   
\definecolor{codeblue}{rgb}{0.0,0.0,0.6}     

\newcommand{\edit}[1]{{\color{black}#1}}

\newcommand{\approach}{{LongCodeZip}\xspace} 

\title{\approach: Compress Long Context for Code Language Models}

\author{\IEEEauthorblockN{Yuling Shi$^{1}$, Yichun Qian$^{2}$, Hongyu Zhang$^{3}$, Beijun Shen$^{1}$, Xiaodong Gu$^{1\ast}$}

\IEEEauthorblockA{$^1$Shanghai Jiao Tong University, Shanghai, China}
\IEEEauthorblockA{$^2$Stanford University, Stanford, CA, USA}
\IEEEauthorblockA{$^3$Chongqing University, Chongqing, China}

\IEEEauthorblockA{\{yuling.shi, bjshen, xiaodong.gu\}@sjtu.edu.cn, ycqian@stanford.edu, hyzhang@cqu.edu.cn}
}

\begin{document}
\maketitle
\thispagestyle{plain}

\begin{abstract}
Code generation under long contexts is becoming increasingly critical as Large Language Models (LLMs) are required to reason over extensive information in the codebase. While recent advances enable code LLMs to process long inputs, high API costs and generation latency remain substantial bottlenecks. Existing context pruning techniques, such as LLMLingua, achieve promising results for general text but overlook code-specific structures and dependencies, leading to suboptimal performance in programming tasks.
In this paper, we propose \approach, a novel plug-and-play code compression framework designed specifically for code LLMs. \approach employs a dual-stage strategy: (1) coarse-grained compression, which identifies and ranks function-level chunks using conditional perplexity with respect to the instruction, retaining only the most relevant functions; and (2) fine-grained compression, which segments retained functions into blocks based on perplexity and selects an optimal subset under an adaptive token budget to maximize relevance.
Evaluations across multiple tasks, including code completion, summarization, and question answering, show that \approach consistently outperforms baseline methods, achieving up to a 5.6× compression ratio without degrading task performance. By effectively reducing context size while preserving essential information, \approach enables LLMs to better scale to real-world, large-scale code scenarios, advancing the efficiency and capability of code intelligence applications\footnote{Our code and data are available at \url{https://github.com/YerbaPage/LongCodeZip}}.
\end{abstract}

\newcommand{\token}{t}  
\newcommand{\importance}{S}  
\newcommand{\budget}{B}  

\section{Introduction}
$\let\thefootnote\relax\footnotetext{* Xiaodong Gu is the corresponding author}$


LLMs specialized for code have revolutionized software development by demonstrating remarkable capabilities in code completion~\cite{guo2023longcoder, liu2023repobench}, translation~\cite{wang2024repotransbench, wang2025evoc2rust}, program synthesis~\cite{zeng2025pruning, zhang2023algo, shi2024code} and program repair~\cite{chen2025swe, li2025swe}. Models like DeepSeek-Coder~\cite{guo2024deepseek}, Qwen2.5-Coder~\cite{hui2024qwen2}, Seed-Coder~\cite{zhang2025seed} can reason over diverse programming languages and significantly enhance productivity.
As code LLMs are increasingly deployed for real-world tasks like repository-level question answering~\cite{liu2024repoqa} and long-context code completion~\cite{guo2023longcoder}, there is a growing demand for handling contexts that span tens of thousands of tokens. This need has motivated efforts to extend LLM context windows~\cite{roziere2023code, hui2024qwen2,zhang2025gam}. However, effective handling of long code contexts remains a central bottleneck.
Three major challenges arise in such long code context scenarios. First, as the input context grows, the quadratic complexity of the transformer attention mechanism~\cite{vaswani2017attention} leads to decreased generation efficiency. At the same time, processing longer inputs with LLMs results in rapidly increasing API costs, especially when pricing models are expensive~\cite{zhang2025drdiff,zhang2025kabb}.
Second, LLMs struggle to identify and utilize relevant content amid lengthy inputs~\cite{liu2023lost, li2023loogle}. Third, even though recent LLMs support extended context windows to 128k tokens, these limits can still be reached when processing large files and long conversation histories, leading to context truncation and degraded outputs~\cite{bogomolov2024long}.

These issues are particularly pronounced in code LLMs. Unlike natural language text, source code is highly structured with complex dependencies spanning across functions, classes, and files. Dependencies between variable declarations, function definitions, and their uses often extend beyond what current context windows can accommodate. As a result, LLMs frequently produce code that fails to compile, violates existing patterns, or ignores critical constraints when the relevant context exceeds their window size~\cite{liu2023your}. Consequently, context compression has emerged as a key demand for enabling long-context code understanding. 

Existing approaches to address long context limitations have notable shortcomings when applied to code. 
General text compression methods like LLMLingua~\cite{jiang2023llmlingua} and Selective Context~\cite{li2023compressing} fail to account for code-specific characteristics and often break code structure. Retrieval-augmented generation (RAG)~\cite{cheng2024xrag} reduce context length by selecting relevant code snippets from the repository context, but it merely rely on text similarities, and may overlook implicit dependencies within the context. Traditional code compressors such as DietCode~\cite{zhang2022diet} and SlimCode~\cite{wang2024natural} improve syntax and structure awareness but are generally limited to function-level pruning or short code examples, leaving compression of long context for code largely unaddressed.

\begin{figure*}[t]
    \centering
    \includegraphics[width=0.75\linewidth]{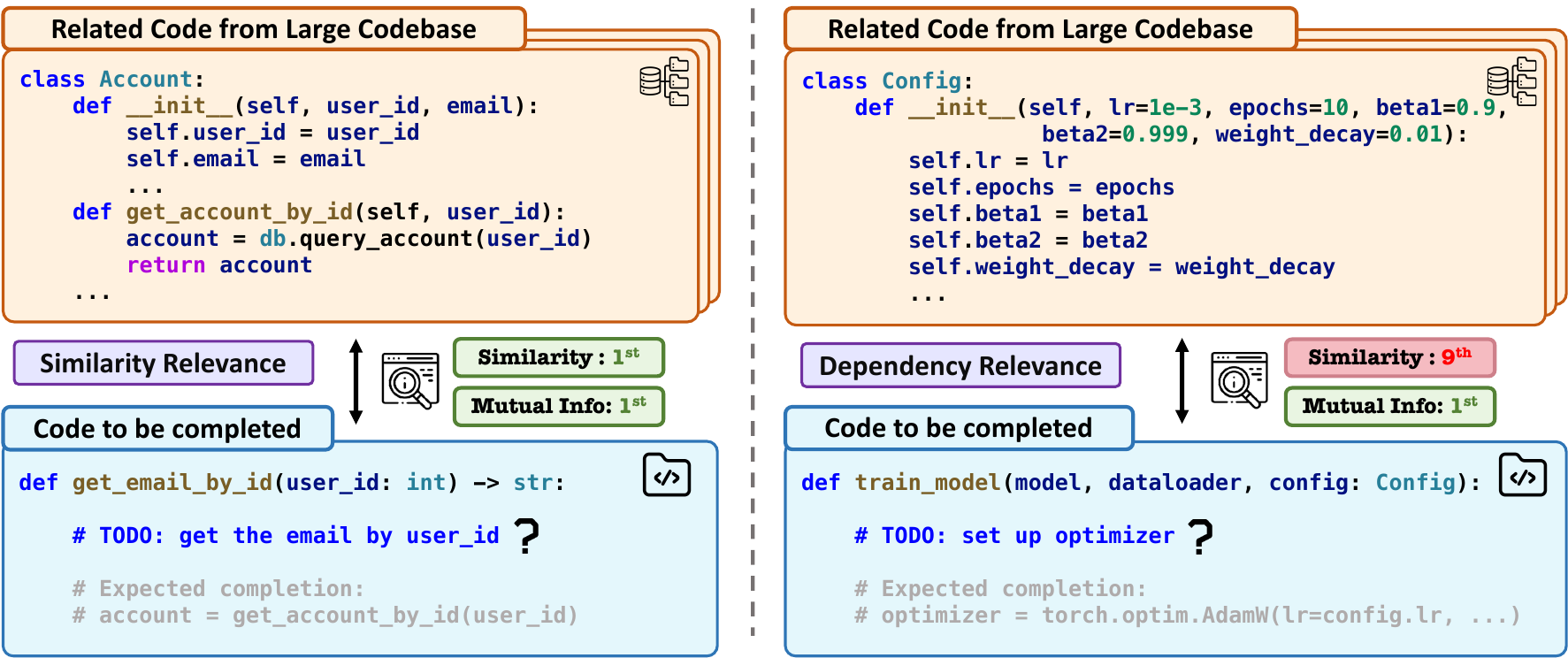}
     \caption{Challenge for RAG, a similariy-based context compression method.}
    \label{fig:motivating_examples}
    \vspace{-0.5cm}
\end{figure*}

To overcome these limitations, we introduce \approach, a training-free, model-agnostic, and plug-and-play context compression framework for code LLMs. Our approach leverages the inherent structure of code through a novel two-stage compression strategy that preserves code semantics while significantly reducing token consumption. First, we perform coarse-grained compression by identifying and ranking function-level chunks based on their relevance to the instruction. Then, within the selected functions, it applies perplexity-based block detection followed by fine-grained block-level compression 
to maximize relevance under an adaptive token budget. To the best of our knowledge, \approach is the first framework specifically designed for long-context code compression and to introduce perplexity-based block detection, providing an efficient and general-purpose solution that preserves task-critical content within strict token limitations. 

We evaluate \approach across multiple code benchmarks with long contexts, including Long Code Completion~\cite{guo2023longcoder}, Long Module Summarization~\cite{bogomolov2024long}, and RepoQA~\cite{liu2024repoqa}. 
Results demonstrate that our approach achieves up to a 5.6× compression ratio without sacrificing performance, generalizes well across tasks and models (even with only 0.5B model as the compressor), and significantly reduces generation time and token costs.

Our main contributions include:
\begin{enumerate}
    \item A novel long-context, code-specific hierarchical compression approach that performs function-level chunking and selection, followed by perplexity-based block detection and block-level pruning.
    \item An adaptive budget allocation and 0/1 knapsack selection mechanism that prioritizes relevant blocks and maximizes critical detail within the token budget.
    \item A comprehensive evaluation demonstrating that \approach outperforms baselines on code completion, summarization, and question answering tasks, achieving up to a 5.6× compression ratio without sacrificing performance.
\end{enumerate}

\section{Motivation}\label{sec:motivation}


Code generation under long context is becoming increasingly important in LLM-based software development. Such tasks often require referencing numerous related files across an entire project repository, resulting in input contexts that span tens of thousands of tokens.
However, these long contexts typically contain scattered and redundant information, which can distract the model and degrade output quality. Moreover, the substantial computational cost of processing such large inputs further exacerbates latency and resource constraints, creating a significant bottleneck for practical deployment.

Retrieval-augmented generation (RAG)~\cite{li2024retrieval,zhang2023repocoder} provides an efficient way to condense overly lengthy contexts. RAG retrieves and appends relevant code snippets to the prompt, leveraging embedding models such as UniXcoder~\cite{guo2022unixcoder} or CodeBERT~\cite{feng2020codebert}, and similarity measures such as cosine similarity. 
While RAG effectively reduces context length, it primarily relies on surface-level lexical similarity between snippets. Consequently, it often fails to capture code segments with deeper semantic or functional dependencies—particularly when such relationships are implicit, abstracted, or span multiple components.


Consider the examples in Figure~\ref{fig:motivating_examples}. In the first scenario, the task is to complete an \texttt{get\_email\_by\_id} function. Retrieving \texttt{Account} class and the \texttt{get\_account\_by\_id} function proves effective, as they share similar function and parameter names. In this case, RAG works well due to strong lexical and structural overlap. 
In the second scenario, however, the task is to implement a \texttt{train\_model} function that relies on configuration values defined in a separate \texttt{Config} class. Here, crucial context like \texttt{Config} is often missed, since RAG may not identify these implicit or non-lexical dependencies. This omission can lead to incomplete or incorrect code generation.


This example highlights the need for context selection criteria that extend beyond surface-level similarity. In both scenarios, an effective similarity measure should assign high relevance to \texttt{get\_account\_by\_id} in the first case and, critically, to \texttt{Config} in the second—even when there is minimal lexical overlap between the configuration class and the training function.


\section{Methodology}
\begin{figure*}[t]
    \centering
    \includegraphics[width=0.7\textwidth]{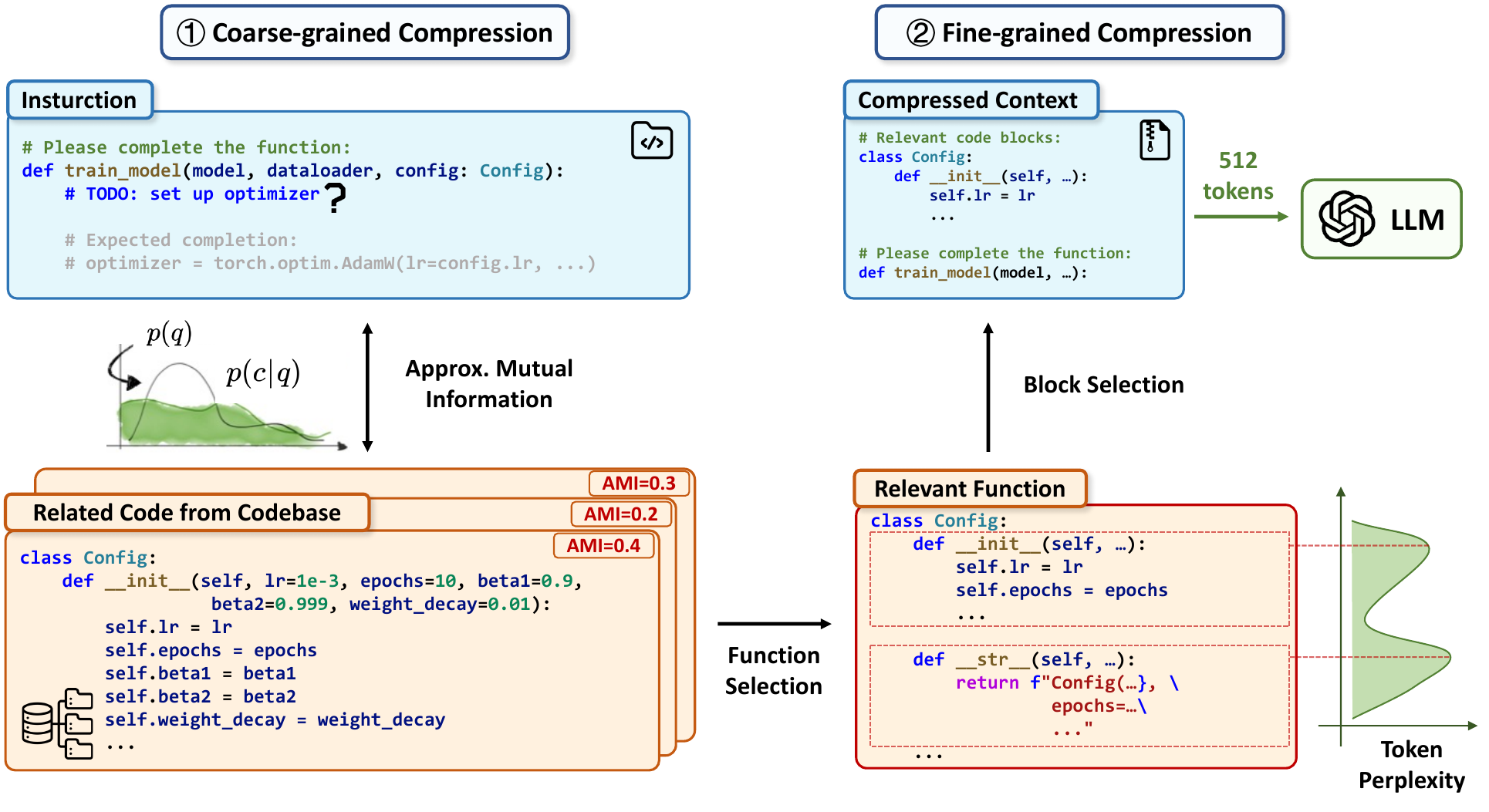}
    \caption{Overview of the \approach framework.} 
    \label{fig:overview}
    \vspace{-1.5em}
\end{figure*}

\subsection{Problem Formulation}

Given a long code context $c = \{c_1, \ldots, c_n\}$ with $n$ tokens and a task instruction $q=\{q_1,...,q_m\}$, the goal of context compression is to produce a compressed context $c' \subseteq c$ such that $|c'| \leq B$, where $B$ is the computational budget in tokens. The objective is to maximize task performance while satisfying the budget constraint. 
For instance, in the code completion task, the instruction could be: \textit{"Complete the following function [code to be completed]"}. The long context could consist of the unfinished code along with retrieved code snippets. 

Rather than relying solely on embedding similarity between $q$ and $c$, we propose to select context snippets based on their mutual information, specifically, how much they reduce the perplexity ($\mathrm{PPL}$) of generating \(q\). Specifically, for each candidate context $c$, we define the approximated mutual information AMI$(c,q)$ as the reduction in perplexity when $c$ is provided:
\begin{equation}
    \text{AMI}(c,q) = \mathrm{PPL}(q) - \mathrm{PPL}(q \mid c)
\label{eq:perplexity_change}
\end{equation}
where $\mathrm{PPL}(q\mid c)$ is the conditional perplexity of $q$ given \(c\), lower values indicate higher likelihood of \(q\)~\cite{jiang2023llmlingua}:
\begin{equation}
\mathrm{PPL}(q | c) = \exp\left( -\frac{1}{N} \sum_{i=1}^{N} \log P(q_i | q_{<i}, c) \right)
\end{equation}
Similarly, $\mathrm{PPL}(q)$ denotes the perplexity of $q$ without the context:
\begin{equation}
\mathrm{PPL}(q) = \exp\left( -\frac{1}{N} \sum_{i=1}^{N} \log P(q_i | q_{<i}) \right)
\label{eq:perplexity}
\end{equation}
Here, \(P\) denotes the model's next-token prediction probability, and $q_{<i}$ is the sequence of preceding tokens before $q_i$. 
A higher AMI score indicates that $c$ enables the model to better predict $q$, capturing both surface-level and dependency-based relevance. We therefore compress long contexts by retaining code snippets with the highest mutual information, ensuring that the most essential information for code generation is preserved.

\subsection{Overview}

The overview of \approach is illustrated in Figure~\ref{fig:overview}. 
Given input of long source code, a task instruction, and a token budget, \approach follows a \emph{coarse-to-fine} compression pipeline. 
In the coarse-grained compression stage (Section \ref{ss:method:coarse}), the source code is divided into function-level chunks, which are ranked by their relevance to the instruction using conditional perplexity. The top \(N\) functions are then selected under a coarse budget, effectively filtering out irrelevant code and avoiding unnecessary computation.
In the fine-grained compression stage (Section \ref{ss:method:finegrained}), each retained function is further segmented into semantic blocks via perplexity-based chunking. An adaptive retention ratio is assigned to each function according to its estimated importance. Within each function, the most relevant blocks are selected by formulating the problem as a 0/1 knapsack optimization, ensuring that the retained content maximizes relevance while fitting within the allocated token budget.

By combining coarse-grained filtering with fine-grained pruning, \approach achieves a balance between aggressive compression and semantic preservation, thereby improving both efficiency and task performance.

\subsection{Coarse-Grained Compression: Relevant Function Selection}
\label{ss:method:coarse}
The coarse-grained compression aims to select high-level code chunks that are most relevant to the task instruction. This process consists of three steps: 

\noindent\textbf{Function-Level Chunking.} We first split the source code into chunks along function or class boundaries. Functions naturally encapsulate coherent logic and exhibit strong modularity~\cite{feng2020codebert}. Chunking at this level ensures that retained code segments are both syntactically valid and semantically self-contained, which is essential for preserving program integrity.

\noindent\textbf{Instruction-aware Relevance Ranking.}
To measure the relevance of each chunk to the task instruction, we employ an instruction-aware ranking mechanism based on approximated mutual information~\eqref{eq:perplexity_change}. Chunks are scored and ranked in descending order, allowing us to prioritize those most informative for the given task.

\noindent\textbf{Budget-Constrained Function Selection.}
Finally, we greedily select the top-ranked chunks under a coarse-grained token budget $B_{\text{coarse}}$, which is the division of the final token budget $B$ by the configurable fine-grained compression ratio $R_{\text{fine}}$. This greedy selection balances efficiency and coverage: a larger budget allows more functions to pass into the fine-grained stage, potentially improving downstream quality but at higher computational cost, while a smaller budget accelerates processing at the risk of discarding useful code. 
Chunks not selected are replaced with placeholders (e.g., comment markers or ellipses), which preserve the global structure while reducing overall context length.


\subsection{Fine-Grained Compression: Intra-Function Pruning}
\label{ss:method:finegrained}

After selecting relevant function-level chunks in the first stage, we apply finer-grained compression to further reduce context length while preserving critical content. This process involves three steps:

\noindent\textbf{Block-Level Chunking.}
The main challenge in intra-function compression is pruning code without breaking internal logic. To address this, each function is segmented into smaller, semantically coherent blocks.
A naive idea is to split code by whitespace lines, but such line-based heuristics often misalign with semantic boundaries. Inspired by techniques in natural language processing~\cite{zhao2024meta}, we employ a perplexity-based method to identify semantic block boundaries within code. While perplexity-based grouping has shown effectiveness in natural language segmentation, it remains under-explored in code. Consecutive lines in code often form strong semantic associations, making perplexity a useful signal. 
Within a semantically coherent region, perplexity tends to decrease as context accumulates~\cite{zhao2024meta}. We treat each line of code as the smallest atomic unit and group consecutive lines based on their perplexity scores, calculated as in \eqref{eq:perplexity}. 
When a line's perplexity exhibits a sharp local increase, exceeding that of its neighbors by at least $\alpha$ times of the standard deviation over all lines, we mark it as a block boundary. Such high-perplexity lines typically mark the beginning of a new block, reflecting underlying semantic or structural changes. This perplexity-guided aggregation allows blocks to capture meaningful code segments while preserving the code structure.



    
        
        

\begin{algorithm}[t]
    \caption{Pseudo code of Adaptive Fine-Grained Budget Allocation}
    \label{algo:adaptive_budget}
    \KwIn{Large functions $\{f_1, ..., f_N\}$ with min-max normalized AMI scores $\{\text{AMI}_1,...,\text{AMI}_N\}$ and token counts $\{T_1, ..., T_N\}$; total token budget for large functions $B_{\text{large}}$; baseline retention ratio $R_{\text{base}}$; importance parameter $\beta$}
    \KwOut{Function-wise adjusted retention rates $\{R_1, ..., R_N\}$}

    $\mathcal{R} \gets \emptyset$\ // Initialize retention rate map \\
  
    \For{$f_i\in\{f_1, ..., f_N\}$}{
        $R_{\text{biased}, i} \gets R_{\text{base}} \cdot (1 + \beta \cdot (2 \times \text{AMI}_{ i} - 1))$; // Compute biased rate
        
        Clamp $R_{\text{biased}, i}$ to $[0, 1]$\;
    }
    \For{$f_i\in \{f_1, ..., f_N\}$}{
        $R_{i} \gets R_{\text{biased}, i} \cdot \frac{B_{\text{large}}}{\sum_{i} R_{\text{biased}, j} \cdot T_j}$;  // Adjust rate\\
    }
    \Return{$R_1, ..., R_N$}\;
\end{algorithm}

\noindent\textbf{Adaptive Budget Allocation.}
Functions selected in the coarse-grained stage vary in importance. Hence, applying a uniform compression ratio across all of them is suboptimal. To address this, we introduce an adaptive budget allocation mechanism that distributes the fine-grained token budget proportionally to function importance. Functions with higher AMI scores receive more token budgets, preserving greater detail, while very small functions $\mathcal{F}_{\text{small}}$ (shorter than five lines) are kept in full. Algorithm~\ref{algo:adaptive_budget} summarizes the procedure. 

We first define the baseline retention ratio for large functions:
\begin{equation}
R_{\text{base}} = \frac{B - \sum_{j \in \mathcal{F}_{\text{small}}} T_j}{\sum_{k \in \mathcal{F}_{\text{large}}} T_k},
\label{eq:global_large_ratio}
\end{equation}
where $B$ is the final token budget, $\mathcal{F}_{\text{small}}$ and $\mathcal{F}_{\text{large}}$ represent the sets of small and large functions respectively, and $T_j$ denotes the number of tokens in function $j$.

For functions \(f_1,\ldots,f_N\) selected in the coarse-grained stage, we perform min-max normalization to all AMI scores to $\text{AMI}_{\text{norm}, i}$.



For each large function $f_i$, and its normalized AMI score $\text{AMI}_{\text{norm},i}\in [0,1]$, a biased retention ratio is then computed as
\begin{equation}
R_{\text{biased}, i} = R_{\text{base}} \cdot (1 + \beta \cdot (2 \times \text{AMI}_{\text{norm}, i} - 1)),
\label{eq:biased_compression}
\end{equation}
where $R_{\text{base}}$ is the baseline retention ratio for large functions (Equation~\ref{eq:global_large_ratio}). The importance parameter $\beta$ adjusts sensitivity to importance. When the importance parameter is set to 0, there is no bias, meaning all functions are treated equally. A more positive $\beta$ increases the emphasis on important functions, allocating more tokens to them. All retention rates are clamped to $[0,1]$ and globally rescaled so that the total number of retained tokens matches the target token budget for large functions $B_{\text{large}}$: 
\begin{equation}
R_{i} = R_{\text{biased},i} \cdot \frac{B_{\text{large}}}{\sum_j R_{\text{biased},j} \cdot T_j},
\label{eq:rate_adjusted}
\end{equation}
where $T_j$ represents the number of tokens in the $j$-th function.
This adjustment preserves the relative importance between functions while ensuring the global constraint is satisfied.

\noindent\textbf{Dynamic Block Selection.}
For each function, \approach identifies a subset of blocks to retain, aiming to maximize the total relevance within the constraints of the allocated token budget. This strategy ensures that the compressed context achieves the highest possible information density. We formulate this selection as a classic 0/1 knapsack problem: each block is treated as an item, where the value corresponds to its normalized AMI score and the weight corresponds to its token length. The detailed procedure is outlined in Algorithm~\ref{algo:knapsack_block_selection}. We employ a dynamic programming approach to compute the optimal subset of blocks that satisfies the budget constraint while maximizing the cumulative value.

    

    

\begin{algorithm}[t]
    \caption{Knapsack Block Selection for Code Compression}
    \label{algo:knapsack_block_selection}
    \KwIn{Blocks $\{b_1, ..., b_N\}$ with min-max normalized AMI scores $\{\text{AMI}_1,...,\text{AMI}_N\}$ and token counts $\{T_1, ..., T_N\}$; token budget of this function $B_i$; user-defined preserved set $\mathcal{P}$}
    \KwOut{Selected blocks $\mathcal{B}_\text{selected} \subseteq \{b_1, ..., b_N\}$}
    
    $B_\text{remain} \gets \max(0, B_i - \sum_{j \in \mathcal{P}} T_j)$\; // Compute remaining budget \\
    \If{$B_\text{remain} = 0$}{
        \Return{$\mathcal{P}$}\;
    }
    $\mathcal{K} \gets \emptyset$\;
    \For{$i = 1$ \KwTo $N$}{
        \If{$i \notin \mathcal{P}$}{
            add $(i, T_i, \text{AMI}_i)$ to $\mathcal{K}$\;
        }
    }

    $\mathcal{B}_\text{selected} \gets \text{0/1 Knapsack DP}(\mathcal{K}, B_\text{remain}) \cup \mathcal{P} $\;
    \Return{$\mathcal{B}_\text{selected}$}\;
\end{algorithm}

    

\begin{table*}[t]
\centering
\caption{Datasets used for evaluating long-context code compression.}
\begin{tabular}{lcccc}
\toprule
\textbf{Dataset} & \textbf{\# Examples} & \textbf{Avg. Context Len.} & \textbf{Avg. Ground Truth Len.} & \textbf{Languages} \\
\midrule
Long Code Completion & 500 & 9,328.2 & 12.4 & Python \\
Long Module Summarization & 139 & 10,809.6 & 1,758.1 & Python \\
Repo QA & 600 & 11,524.6 & 156.0 & Python, Java, JS, Rust, Go, C++ \\
\bottomrule
\end{tabular}
\label{tab:statistics}
\vspace{-1.5em}
\end{table*}

\section{Experimental Setup}

\subsection{Research Questions (RQs)}

\textbf{RQ1:} Can \approach effectively compress code context while preserving the downstream performance?

\textbf{RQ2:} How does different parts of \approach contribute to the performance?

\textbf{RQ3:} Does \approach exhibit cross-model generalization capabilities?

\textbf{RQ4:} What is the efficiency benefit of \approach in downstream tasks?


\subsection{Datasets}

We evaluate our method on long code context benchmarks across three common tasks: code completion, code summarization, and code question answering. These tasks reflect practical developer needs and assess whether compressed code retains sufficient information for downstream performance. For each task, we construct prompts following the benchmark papers~\cite{guo2023longcoder, bogomolov2024long, liu2024repoqa}. Dataset statistics are shown in Table~\ref{tab:statistics}.


The Long Code Completion dataset~\cite{guo2023longcoder} targets the code completion task under long-context of relevant functions. To highlight long-context difficulties, 
we filtered the test set to 500 Python examples with input contexts longer than 5,000 tokens.
The Long Module Summarization dataset~\cite{bogomolov2024long} contains 216 examples from 43 Python repositories. To focus on the challenging long-context scenario, we also further filtered the original dataset to 139 examples that have more than 2,000 context tokens.
RepoQA~\cite{liu2024repoqa} is a multilingual benchmark that contains 600 long code question answering tests across 60 repositories and 6 programming languages. It requires the model to locate and return a function within the long context using a natural language instruction, similar to a retrieval task.




\subsection{Baselines and Models}

We evaluate \approach against a variety of competitive baselines: 

\noindent 1) \textbf{No Compression}: The full code context is used without any compression, representing the upper bound of performance. 

\noindent 2) \textbf{No Context}: The model is evaluated with only the task instruction, without any context, representing the lower bound. 

\noindent 3) \textbf{Random Baselines}: Random Token randomly removes individual tokens, while Random Line randomly removes the whole lines of code. 

\noindent 4) \textbf{Retrieval-based Methods}: RAG (Sliding Window) uses fixed-size overlapping chunks, whereas RAG (Function Chunking) splits code at function boundaries. Both methods use the state-of-the-art code embedding model UniXCoder-base~\cite{guo2022unixcoder,zhang2023repocoder}. 

\noindent 5) \textbf{Code Compression Methods}: We compare against the compression components from DietCode~\cite{zhang2022diet} and SlimCode~\cite{wang2024natural}. DietCode was originally implemented for Python and Java, while SlimCode supports only Java. To enable direct comparison on other benchmarks, we reproduce the SlimCode for Python with tree-sitter\footnote{\url{https://tree-sitter.github.io/tree-sitter/}}. 

\noindent 6) \textbf{Text Compression Methods}: We also include several state-of-the-art prompt compression methods for natural languages, including LLMLingua~\cite{jiang2023llmlingua}, LongLLMLingua~\cite{jiang2024longllmlingua}, and LLMLingua-2~\cite{pan2024llmlingua}. 

All methods are evaluated on a diverse set of code LLMs, covering both earlier models like Deepseek-Coder-6.7B~\cite{guo2024deepseek} and latest models like Qwen2.5-Coder-7B~\cite{hui2024qwen2} and Seed-Coder-8B~\cite{zhang2025seed}. Specifically, we use the instruct version of these models from Huggingface\footnote{\url{https://huggingface.co/models}}. \edit{To further demonstrate the generalizability of \approach, we also extend our evaluation to state-of-the-art closed-source models, including GPT-4o\footnote{\url{https://openai.com/index/hello-gpt-4o/}} and Claude-3.7-Sonnet\footnote{\url{https://www.anthropic.com/news/claude-3-7-sonnet}}.}

\subsection{Evaluation Metrics}
The evaluation of \approach encompasses two primary dimensions: compression efficiency and downstream generation performance with compressed context.
We report compression \textit{Ratio} on all tasks:
\begin{equation}
\textit{Ratio} = \frac{|C_{\text{original}}|}{|C_{\text{compressed}}|}
\end{equation}
where $|C_{\text{compressed}}|$ and $|C_{\text{original}}|$ denote the number of tokens in the compressed and original contexts respectively.

For the code completion task, We follow LongCoder~\cite{guo2023longcoder} to evaluate the performance of the models in terms of Exact Match (EM) and Edit Similarity (ES). 

For the code summarization task, we follow~\cite{bogomolov2024long} to use a third party model \textsc{GPT-4o-mini}~\footnote{\url{https://platform.openai.com/docs/models/gpt-4o-mini}} to evaluate which summary better explains the code between the ground truth and the generated one. \edit{This LLM-as-Judge evaluation strategy is widely adopted in both NLP and software engineering domains~\cite{liu2023g,song2024finesure}, as it has been demonstrated to align well with human preferences and provides more nuanced evaluation compared to traditional metrics~\cite{wang2025can,he2025code}.} The model chooses the better summary after reviewing both options alongside the code. To avoid bias, we prompt it twice with the order reversed. We then compute \textit{CompScore} as:
\begin{equation}
\textit{CompScore} = \frac{1}{2}[\mathcal{P}(s_o \succ \hat{s}) + (1 - \mathcal{P}(\hat{s} \succ s_o))]
\end{equation}
where $\mathcal{P}(s_o \succ \hat{s})$ is the probability that the referee model prefers the generated summary $s_o$ over the reference $\hat{s}$, and $\mathcal{P}(\hat{s} \succ s_o)$ is the probability for the reverse order. $\mathcal{S}_{\text{comp}}$ ranges from 0 to 100, with 50 indicating equal preference.

For the code QA task, we follow~\cite{liu2024repoqa} to evaluate the retrieval accuracy of needle functions, reporting the percentage of models that retrieve a correct match above a BLEU similarity threshold of 0.8 between the generated function $f_o$ and the target function $\hat{f}$: $\text{BLEU}(\hat{f}, f_o) > 0.8$.

\subsection{Implementation Details}
We tailored hyperparameters for each distinct task, consistently mirroring the generation model with our compression model. For \textbf{code completion}, which demands focused context, we set the token budget $B$ to $2 \text{k}$, the fine-grained ratio ($R_{\text{fine}}$) to $0.8$, and the importance adjustment parameter ($\beta$) to $0.5$. Conversely, \textbf{code summarization} necessitates understanding broader context within large modules; consequently, we increased $B$ to $5 \text{k}$, reduced $R_{\text{fine}}$ to $0.3$, and maintained $\beta$ at $0.5$. For the \textbf{RepoQA task}, where the objective is to precisely replicate entire functions, we set $B$ to $2 \text{k}$ and $R_{\text{fine}}$ to $1.0$ to ensure the structural integrity of functions. These values for $B$, $\beta$, and $R_{\text{fine}}$ were determined through experiments on a small held-out set that did not overlap with the test data. All experiments were conducted on a system equipped with an Intel Xeon Gold 6254 CPU and an NVIDIA A100-80G GPU.

\section{Results}

\subsection{RQ1: Effectiveness on Code Compression}

\begin{table}[t]
    \centering
    \caption{Results on Long Code Completion}
    \resizebox{\columnwidth}{!}{
    \begin{tabular}{llccc}
    \toprule
    \textbf{Model} & \textbf{Method} & \textbf{ES} & \textbf{EM} & \textbf{Ratio} \\
    \midrule
    \multirow{13}{*}{\textsc{Deepseek-Coder-6.7B}} & \textit{No Compression} & 57.14 & 34.40 & 1.0x \\
    & \textit{No Context} & 41.29 & 13.20 & - \\
    \cmidrule{2-5}
    & \textit{Random Token} & 44.86 & 13.40 & 4.4x \\
    & \textit{Random Line} & 50.54 & 21.20 & 4.5x \\
    \cmidrule{2-5}
    & RAG (Sliding Window) & 58.48 & 31.60 & 4.2x \\
    & RAG (Function Chunking) & 57.93 & 30.80 & 5.7x \\
    & LLMLingua & 43.61 & 14.00 & 5.6x \\
    & LLMLingua-2 & 46.23 & 15.00 & 4.4x \\
    & LongLLMLingua & 54.09 & 26.40 & 4.8x \\
    & DietCode & 51.57 & 20.20 & 3.4x \\
    & SlimCode & 48.84 & 19.80 & 4.5x \\
    \cmidrule(lr){2-5}
    & \cellcolor{cyan!10}\textbf{\approach} & \cellcolor{cyan!10}\textbf{60.58} & \cellcolor{cyan!10}\textbf{35.40} & \cellcolor{cyan!10}5.3x \\
    \midrule
    \multirow{13}{*}{\textsc{Qwen2.5-Coder-7B}} & \textit{No Compression} & 56.36 & 31.80 & 1.0x \\
    & \textit{No Context} & 38.14 & 9.60 & - \\
    \cmidrule{2-5}
    & \textit{Random Token} & 39.10 & 8.40 & 4.4x \\
    & \textit{Random Line} & 39.73 & 12.40 & 4.5x \\
    \cmidrule{2-5}
    & RAG (Sliding Window) & 50.81 & 24.60 & 2.8x \\
    & RAG (Function Chunking) & 52.79 & 26.00 & 3.1x \\
    & LLMLingua & 21.56 & 5.40 & 3.4x \\
    & LLMLingua-2 & 41.29 & 12.20 & 4.4x \\
    & LongLLMLingua & 23.88 & 9.00 & 3.2x \\
    & DietCode & 43.91 & 13.20 & 3.4x \\
    & SlimCode & 40.85 & 12.20 & 4.5x \\
    \cmidrule(lr){2-5}
    & \cellcolor{cyan!10}\textbf{\approach} & \cellcolor{cyan!10}\textbf{57.55} & \cellcolor{cyan!10}\textbf{32.40} & \cellcolor{cyan!10}4.3x \\
    \midrule
    \multirow{13}{*}{\textsc{Seed-Coder-8B}} & \textit{No Compression} & 64.04 & 40.20 & 1.0x \\
    & \textit{No Context} & 41.88 & 13.60 & - \\
    \cmidrule{2-5}
    & \textit{Random Token} & 45.35 & 13.40 & 4.4x \\
    & \textit{Random Line} & 50.10 & 21.20 & 4.5x \\
    \cmidrule{2-5}
    & RAG (Sliding Window) & 58.51 & 32.40 & 2.8x \\
    & RAG (Function Chunking) & 60.52 & 35.00 & 3.7x \\
    & LLMLingua & 44.36 & 14.40 & 4.5x \\
    & LLMLingua-2 & 46.69 & 15.40 & 4.4x \\
    & LongLLMLingua & 54.84 & 26.40 & 4.2x \\
    & DietCode & 51.43 & 18.80 & 3.4x \\
    & SlimCode & 50.45 & 19.80 & 4.5x \\
    \cmidrule(lr){2-5}
    & \cellcolor{cyan!10}\textbf{\approach} & \cellcolor{cyan!10}\textbf{63.11} & \cellcolor{cyan!10}\textbf{37.40} & \cellcolor{cyan!10}5.6x \\
    \bottomrule
    \end{tabular}
    }
    \label{tab:lcc_results}
\end{table}

\begin{table}[t]
    \centering
    \caption{Results on Long Module Summarization}
    \resizebox{0.97\columnwidth}{!}{
    \label{tab:lms_results}
    \begin{tabular}{llcc}
    \toprule
    \textbf{Model} & \textbf{Method} & \textbf{CompScore} & \textbf{Ratio} \\
    \midrule
    \multirow{13}{*}{\textsc{Deepseek-Coder-6.7B}} & \textit{No Compression} & 19.09 & 1.0x \\
    & \textit{No Context} & 2.49 & - \\
    \cmidrule{2-4}
    & \textit{Random Token} & 11.88 & 1.8x \\
    & \textit{Random Line} & 17.62 & 1.8x \\
    \cmidrule{2-4}
    & RAG (Sliding Window) & 22.95 & 2.1x \\
    & RAG (Function Chunking) & 18.47 & 2.1x \\
    & LLMLingua & 17.65 & 2.1x \\
    & LongLLMLingua & 21.62 & 1.7x \\
    & LLMLingua-2 & 18.48 & 2.1x \\
    & DietCode & 17.35 & 2.1x \\
    & SlimCode & 20.24 & 2.2x \\
    \cmidrule(lr){2-4}
    & \cellcolor{cyan!10}\textbf{\approach} & \cellcolor{cyan!10}\textbf{28.01} & \cellcolor{cyan!10}2.5x \\
    \midrule
    \multirow{13}{*}{\textsc{Qwen2.5-Coder-7B}} & \textit{No Compression} & 56.00 & 1.0x \\
    & \textit{No Context} & 6.13 & - \\
    \cmidrule{2-4}
    & \textit{Random Token} & 34.09 & 1.8x \\
    & \textit{Random Line} & 46.19 & 1.8x \\
    \cmidrule{2-4}
    & RAG (Sliding Window) & 53.50 & 1.7x \\
    & RAG (Function Chunking) & 40.84 & 2.1x \\
    & LLMLingua & 39.81 & 1.7x \\
    & LongLLMLingua & 46.72 & 1.5x \\
    & LLMLingua-2 & 52.99 & 2.1x \\
    & DietCode & 35.67 & 2.1x \\
    & SlimCode & 44.13 & 2.2x \\
    \cmidrule(lr){2-4}
    & \cellcolor{cyan!10}\textbf{\approach} & \cellcolor{cyan!10}\textbf{56.47} & \cellcolor{cyan!10}1.7x \\
    \midrule
    \multirow{13}{*}{\textsc{Seed-Coder-8B}} & \textit{No Compression} & 44.95 & 1.0x \\
    & \textit{No Context} & 17.42 & - \\
    \cmidrule{2-4}
    & \textit{Random Token} & 34.16 & 1.8x \\
    & \textit{Random Line} & 41.27 & 1.8x \\
    \cmidrule{2-4}
    & RAG (Sliding Window) & 42.54 & 3.0x \\
    & RAG (Function Chunking) & 43.19 & 2.1x \\
    & LLMLingua & 32.00 & 3.1x \\
    & LongLLMLingua & 49.73 & 2.4x \\
    & LLMLingua-2 & 53.88 & 3.2x \\
    & DietCode & 44.74 & 2.1x \\
    & SlimCode & 46.01 & 2.2x \\
    \cmidrule(lr){2-4}
    & \cellcolor{cyan!10}\textbf{\approach} & \cellcolor{cyan!10}\textbf{55.07} & \cellcolor{cyan!10}3.5x \\
    \bottomrule
    \end{tabular}
    \vspace{-1cm}
    }
\end{table}

\begin{table}[h] 
    \centering
    \caption{Results on RepoQA}
    \label{tab:repoqa_results} 
    \setlength{\tabcolsep}{4pt} 
    \resizebox{0.95\columnwidth}{!}{%
    \begin{tabular}{l@{\hspace{2pt}}cccccc|c@{\hspace{3pt}}c} 
    \toprule
    \textbf{Method} & \textbf{Py} & \textbf{C++} & \textbf{Java} & \textbf{TS} & \textbf{Rust} & \textbf{Go} & \textbf{Avg.} & \textbf{Ratio} \\
    \midrule
    \multicolumn{9}{>{\columncolor{gray!20}}c}{\textbf{\textsc{Deepseek-Coder-6.7b}}} \\ 
    \cmidrule(lr){1-9} 
    \textit{No Compression} & 21.0 & 30.0 & 44.0 & 49.0 & 27.0 & 59.0 & 38.3 & 1.0x \\
    \textit{No Context} & 0.0 & 0.0 & 0.0 & 0.0 & 0.0 & 0.0 & 0.0 & - \\
    \cmidrule{1-9} 
    \textit{Random Token} & 0.0 & 1.0 & 2.0 & 1.0 & 0.0 & 6.0 & 1.7 & 3.6x \\
    \textit{Random Line} & 3.0 & 12.0 & 9.0 & 7.0 & 5.0 & 8.0 & 7.3 & 3.5x \\
    \cmidrule{1-9} 
    RAG (Sliding Window) & 49.0 & 55.0 & 53.0 & 67.0 & 47.0 & 62.0 & 55.5 & 3.5x \\
    RAG (Function Chunking) & 42.0 & 40.0 & 30.0 & 36.0 & 49.0 & 57.0 & 42.3 & 4.0x \\
    LLMLingua & 0.0 & 2.0 & 6.0 & 1.0 & 2.0 & 4.0 & 2.5 & 3.6x \\
    LLMLingua-2 & 1.0 & 1.0 & 4.0 & 0.0 & 0.0 & 3.0 & 1.5 & 4.6x \\
    LongLLMLingua & 52.0 & 54.0 & 65.0 & 62.0 & 56.0 & 67.0 & 59.3 & 3.0x \\
    DietCode & 13.0 & - & 28.0 & - & - & - & 20.5 & 3.7x\\
    SlimCode & 15.0 & - & 35.0 & - & - & - & 25.0 & 4.3x\\
    \cmidrule(lr){1-9} 
    \cellcolor{cyan!10}\textbf{\approach} & \cellcolor{cyan!10}\textbf{76.0} & \cellcolor{cyan!10}\textbf{69.0} & \cellcolor{cyan!10}\textbf{80.0} & \cellcolor{cyan!10}\textbf{75.0} & \cellcolor{cyan!10}\textbf{73.0} & \cellcolor{cyan!10}\textbf{79.0} & \cellcolor{cyan!10}\textbf{75.3} & \cellcolor{cyan!10}5.3x \\
    \midrule

    \multicolumn{9}{>{\columncolor{gray!20}}c}{\textbf{\textsc{Qwen2.5-Coder-7B}}} \\
    \cmidrule(lr){1-9}
    \textit{No Compression} & 84.0 & 77.0 & 89.0 & 93.0 & 83.0 & 90.0 & 86.0 & 1.0x \\
    \textit{No Context} & 0.0 & 0.0 & 0.0 & 0.0 & 0.0 & 0.0 & 0.0 & - \\
    \cmidrule{1-9}
    \textit{Random Token} & 1.0 & 3.0 & 4.0 & 2.0 & 4.0 & 7.0 & 3.5 & 3.6x \\
    \textit{Random Line} & 6.0 & 11.0 & 22.0 & 10.0 & 9.0 & 13.0 & 11.8 & 3.5x \\
    \cmidrule{1-9}
    RAG (Sliding Window) & 64.0 & 65.0 & 68.0 & 72.0 & 57.0 & 79.0 & 67.5 & 3.7x \\
    RAG (Function Chunking) & 54.0 & 47.0 & 59.0 & 39.0 & 58.0 & 69.0 & 54.3 & 4.3x \\
    LLMLingua & 5.0 & 7.0 & 9.0 & 11.0 & 4.0 & 16.0 & 8.7 & 4.1x \\
    LLMLingua-2 & 1.0 & 2.0 & 8.0 & 1.0 & 1.0 & 4.0 & 2.8 & 4.6x \\
    LongLLMLingua & 70.0 & 63.0 & 71.0 & 68.0 & 78.0 & 78.0 & 71.3 & 4.3x \\
    DietCode & 17.0 & - & 35.0 & - & - & - & 26.0 & 3.7x\\
    SlimCode & 20.0 & - & 48.0 & - & - & - & 34.0 & 4.3x\\
    \cmidrule(lr){1-9}
    \cellcolor{cyan!10}\textbf{\approach} & \cellcolor{cyan!10}\textbf{92.0} & \cellcolor{cyan!10}\textbf{78.0} & \cellcolor{cyan!10}\textbf{87.0} & \cellcolor{cyan!10}\textbf{85.0} & \cellcolor{cyan!10}\textbf{86.0} & \cellcolor{cyan!10}\textbf{95.0} & \cellcolor{cyan!10}\textbf{87.2} & \cellcolor{cyan!10}4.5x \\
    \midrule

    \multicolumn{9}{>{\columncolor{gray!20}}c}{\textbf{\textsc{Seed-Coder-8B}}} \\
    \cmidrule(lr){1-9}
    \textit{No Compression} & 73.0 & 52.0 & 70.0 & 81.0 & 57.0 & 81.0 & 69.0 & 1.0x \\
    \textit{No Context} & 0.0 & 0.0 & 0.0 & 0.0 & 0.0 & 0.0 & 0.0 & - \\
    \cmidrule{1-9}
    \textit{Random Token} & 2.0 & 3.0 & 4.0 & 1.0 & 1.0 & 10.0 & 3.5 & 3.6x \\
    \textit{Random Line} & 5.0 & 6.0 & 17.0 & 6.0 & 4.0 & 18.0 & 9.3 & 3.5x \\
    \cmidrule{1-9}
    RAG (Sliding Window) & 58.0 & 51.0 & 66.0 & 64.0 & 57.0 & 74.0 & 61.7 & 3.9x \\
    RAG (Function Chunking) & 49.0 & 40.0 & 50.0 & 30.0 & 47.0 & 64.0 & 46.7 & 4.5x \\
    LLMLingua & 4.0 & 3.0 & 9.0 & 8.0 & 5.0 & 10.0 & 6.5 & 4.3x \\
    LLMLingua-2 & 1.0 & 2.0 & 4.0 & 1.0 & 1.0 & 6.0 & 2.5 & 4.6x \\
    LongLLMLingua & 71.0 & 60.0 & 74.0 & 65.0 & 74.0 & 83.0 & 71.2 & 5.1x \\
    DietCode & 16.0 & - & 32.0 & - & - & - & 24.0 & 3.7x\\
    SlimCode & 25.0 & - & 50.0 & - & - & - & 37.5 & 4.3x\\
    \cmidrule(lr){1-9}
    \cellcolor{cyan!10}\textbf{\approach} & \cellcolor{cyan!10}\textbf{83.0} & \cellcolor{cyan!10}\textbf{70.0} & \cellcolor{cyan!10}\textbf{92.0} & \cellcolor{cyan!10}\textbf{74.0} & \cellcolor{cyan!10}\textbf{78.0} & \cellcolor{cyan!10}\textbf{87.0} & \cellcolor{cyan!10}\textbf{80.7} & \cellcolor{cyan!10}5.3x \\
    \midrule
    \bottomrule
    \end{tabular}
    \vspace{-1cm}
    }
\end{table}

Tables~\ref{tab:lcc_results}, \ref{tab:lms_results}, and \ref{tab:repoqa_results} present the evaluation results of our approach on three downstream tasks, respectively. The best scores among compression methods are bolded. Across all three tasks and multiple backbone models, \approach consistently outperforms compression baselines by substantial and statistically significant margins ($p < 0.001$ via Wilcoxon signed-rank test on 10 repeated experiments), even when operating at comparable or stricter compression ratios.

\begin{table*}[t]
    \centering
    \caption{\edit{Results with Closed-source Models}}
    \label{tab:claude_gpt4o_comprehensive}
    \begin{threeparttable}
    \setlength{\tabcolsep}{4pt}
    \resizebox{0.9\textwidth}{!}{%
    \begin{tabular}{l|ccc|ccc|cc|cc|cc|cc}
    \toprule
    & \multicolumn{6}{c|}{\textbf{Long Code Completion}} & \multicolumn{4}{c|}{\textbf{Long Module Summarization}} & \multicolumn{4}{c}{\textbf{RepoQA}} \\
    \cmidrule(lr){2-7} \cmidrule(lr){8-11} \cmidrule(lr){12-15}
    \multirow{2}{*}{\textbf{Method}} & \multicolumn{3}{c|}{\textbf{\textsc{Claude-3.7-Sonnet}}} & \multicolumn{3}{c|}{\textbf{\textsc{GPT-4o}}} & \multicolumn{2}{c|}{\textbf{\textsc{Claude-3.7-Sonnet}}} & \multicolumn{2}{c|}{\textbf{\textsc{GPT-4o}}} & \multicolumn{2}{c|}{\textbf{\textsc{Claude-3.7-Sonnet}}} & \multicolumn{2}{c}{\textbf{\textsc{GPT-4o}}} \\
    \cmidrule(lr){2-4} \cmidrule(lr){5-7} \cmidrule(lr){8-9} \cmidrule(lr){10-11} \cmidrule(lr){12-13} \cmidrule(lr){14-15}
    & \textbf{ES} & \textbf{EM} & \textbf{Ratio} & \textbf{ES} & \textbf{EM} & \textbf{Ratio} & \textbf{CompScore} & \textbf{Ratio} & \textbf{CompScore} & \textbf{Ratio} & \textbf{Avg Acc} & \textbf{Ratio} & \textbf{Avg Acc} & \textbf{Ratio} \\
    \midrule
    \textit{No Compression} & 66.24 & 41.20 & 1.0x & 65.13 & 40.80 & 1.0x & 60.72 & 1.0x & 58.42 & 1.0x & 89.7 & 1.0x & 87.8 & 1.0x \\
    \textit{No Context} & 43.97 & 14.20 & - & 42.92 & 14.00 & - & 6.58 & - & 6.41 & - & 0.0 & - & 0.0 & - \\
    \cmidrule{1-15}
    \textit{Random Token} & 47.61 & 14.00 & 4.4x & 46.51 & 13.80 & 4.4x & 37.45 & 1.8x & 35.83 & 1.8x & 3.8 & 3.6x & 3.8 & 3.6x \\
    \textit{Random Line} & 52.61 & 22.20 & 4.5x & 51.42 & 21.80 & 4.5x & 50.12 & 1.8x & 48.24 & 1.8x & 12.2 & 3.5x & 12.1 & 3.5x \\
    \cmidrule{1-15}
    RAG (Sliding Window) & 61.44 & 34.00 & 2.8x & 60.03 & 33.20 & 2.8x & 58.03 & 1.7x & 55.85 & 1.7x & 73.8 & 3.7x & 73.0 & 3.7x \\
    RAG (Function Chunking) & 63.55 & 36.80 & 3.1x & 62.01 & 36.00 & 3.1x & 44.56 & 2.1x & 42.76 & 2.1x & 55.0 & 4.3x & 52.5 & 4.3x \\
    LLMLingua & 46.58 & 15.20 & 3.4x & 45.53 & 14.80 & 3.4x & 43.21 & 1.7x & 41.57 & 1.7x & 2.8 & 4.1x & 2.7 & 4.1x \\
    LLMLingua-2 & 49.02 & 16.20 & 4.4x & 47.90 & 15.80 & 4.4x & 57.85 & 2.1x & 55.48 & 2.1x & 3.0 & 4.6x & 2.8 & 4.6x \\
    LongLLMLingua & 57.58 & 27.80 & 3.2x & 56.24 & 27.20 & 3.2x & 50.86 & 1.5x & 48.89 & 1.5x & 74.5 & 4.8x & 73.2 & 4.8x \\
    DietCode & 54.00 & 19.80 & 3.4x & 52.76 & 19.40 & 3.4x & 38.82 & 2.1x & 37.21 & 2.1x & 26.7 & 3.7x & 25.5 & 3.7x \\
    SlimCode & 53.03 & 20.80 & 4.5x & 51.78 & 20.40 & 4.5x & 48.11 & 2.2x & 46.13 & 2.2x & 38.3 & 4.3x & 37.0 & 4.3x \\
    \cmidrule(lr){1-15}
    \cellcolor{cyan!10}\textbf{\approach} & \cellcolor{cyan!10}\textbf{66.27} & \cellcolor{cyan!10}\textbf{40.20} & \cellcolor{cyan!10}4.3x & \cellcolor{cyan!10}\textbf{64.72} & \cellcolor{cyan!10}\textbf{38.80} & \cellcolor{cyan!10}4.3x & \cellcolor{cyan!10}\textbf{61.47} & \cellcolor{cyan!10}1.7x & \cellcolor{cyan!10}\textbf{59.04} & \cellcolor{cyan!10}1.7x & \cellcolor{cyan!10}\textbf{88.9} & \cellcolor{cyan!10}5.1x & \cellcolor{cyan!10}\textbf{88.9} & \cellcolor{cyan!10}5.1x \\
    \bottomrule
    \end{tabular}
    }
    \end{threeparttable}
    \vspace{-0.25cm}
\end{table*}

Specifically, on the Long Code Completion task, RAG-based methods achieve higher ES and EM scores than other baseline methods, but still fall short of our approach. For instance, with Qwen2.5-Coder-7B, RAG (Function Chunking) achieves an ES score of 52.79 and an EM score of 26.00 at a 3.1$\times$ compression ratio. In contrast, our approach achieves 57.55 ES and 32.40 EM at a stricter 4.3$\times$ compression ratio, representing a 28\% shorter compressed context than the RAG method. This demonstrates that our method not only preserves more critical information for code completion but also does so with significantly greater compression efficiency.

In contrast to the code completion results, RAG-based methods do not show clear advantages over other baselines on the Long Module Summarization task. However, our approach remains the most competitive, achieving a \textit{CompScore} of 28.01 with Deepseek-Coder-6.7B at a 2.5$\times$ compression ratio—surpassing other compression baselines by a considerable margin. 
This highlights the effectiveness of our method in preserving relevant semantic content for summarization, even with shorter input contexts.

On the RepoQA task, LLMLingua and LLMLingua-2 exhibit poor performance because token-level compression corrupts code syntax and structure, while LongLLMLingua improves this dramatically by performing coarse-grained document-level to fine-grained token-level compression, using instruction-aware contrastive perplexity to preserve code segments highly relevant to the instruction. Nonetheless, our approach consistently achieves the best performance across all models. Notably, on Deepseek-Coder-6.7B, our approach surpasses LongLLMLingua by 16\% in overall score while compressing the context to half the length. This underscores the superior effectiveness of our method in both information retention and aggressive compression for long code understanding.

\edit{Notably, \approach demonstrates strong generalizability across state-of-the-art closed-source models. As comprehensively shown in Table~\ref{tab:claude_gpt4o_comprehensive}, on GPT-4o, \approach achieves an ES score of 64.72 (vs. 65.13 no-compression baseline) on Long Code Completion at a 4.3x compression ratio, closely matching the performance of the uncompressed input while significantly reducing context length. For the RepoQA task, \approach even surpasses the no-compression baseline, achieving 88.9 average score on GPT-4o, demonstrating that removing irrelevant context can improve performance on complex reasoning tasks. On the more powerful Claude-3.7-Sonnet, \approach achieves 66.27 ES (vs. 66.24 baseline) with the same compression efficiency. For the RepoQA task, \approach also surpasses the no-compression baseline on Claude-3.7-Sonnet, achieving 90.7 average score, further demonstrating the effectiveness of our approach.} 

\edit{We also conduct comprehensive comparisons with recent advanced approaches in code completion, including A\textsuperscript{3}-CodGen~\cite{liao2024a3codgen}, cAST~\cite{zhang2025cast}, RepoGenix~\cite{liang2024repogenix}, and RLCoder~\cite{wang2024rlcoder} across all evaluated models. As shown in Table~\ref{tab:advanced_rag_comparison}, \approach consistently outperforms these advanced RAG methods across the most competitive open-source and closed-source models, SeedCoder and Claude-3.7-Sonnet. Our method can more efficiently retain essential information, achieving higher information density under the same token budget. 
This demonstrates the broad applicability and consistent effectiveness of our approach across diverse model architectures and capabilities. Notably, these RAG-based retrieval methods are complementary to our compression approach and could potentially be combined with our framework to further enhance performance by first retrieving relevant content and then applying our compression techniques.}

Overall, our method achieves effectiveness on par with or better than the No Compression setting, and consistently outperforms all compression baselines across tasks and backbone models even under more aggressive compression.

\begin{finding}{Finding 1}
\approach is effective across various downstream tasks, with up to 5.6x compression ratio without sacrificing downstream performance.
\end{finding}

\begin{table}[t]
    \centering
    \caption{\edit{Comparison with Advanced RAG Methods on Long Code Completion}}
    \label{tab:advanced_rag_comparison}
    \resizebox{\columnwidth}{!}{
    \begin{tabular}{llccc}
    \toprule
    \textbf{Model} & \textbf{Method} & \textbf{ES} & \textbf{EM} & \textbf{Ratio} \\
    \midrule
    \multirow{6}{*}{\textsc{Seed-Coder-8B}} & \textit{No Compression} & 64.04 & 40.20 & 1.0x \\
    \cmidrule{2-5}
    & A\textsuperscript{3}-CodGen & 58.70 & 33.10 & 3.8x \\
    & cAST & 57.35 & 30.90 & 4.1x \\
    & RepoGenix & 60.28 & 34.70 & 3.5x \\
    & RLCoder & 58.14 & 32.30 & 4.0x \\
    \cmidrule(lr){2-5}
    & \cellcolor{cyan!10}\textbf{\approach} & \cellcolor{cyan!10}\textbf{63.11} & \cellcolor{cyan!10}\textbf{37.40} & \cellcolor{cyan!10}5.6x \\
    \midrule
    \multirow{6}{*}{\textsc{Claude-3.7-Sonnet}} & \textit{No Compression} & 66.24 & 41.20 & 1.0x \\
    \cmidrule{2-5}
    & A\textsuperscript{3}-CodGen & 60.15 & 35.80 & 3.8x \\
    & cAST & 58.92 & 33.60 & 4.1x \\
    & RepoGenix & 62.48 & 37.40 & 3.5x \\
    & RLCoder & 62.76 & 37.90 & 4.0x \\
    \cmidrule(lr){2-5}
    & \cellcolor{cyan!10}\textbf{\approach} & \cellcolor{cyan!10}\textbf{66.27} & \cellcolor{cyan!10}\textbf{40.20} & \cellcolor{cyan!10}4.3x \\
    \bottomrule
    \end{tabular}
    }
\end{table}

\begin{table}[t]
    \centering
    \caption{Ablation Study Results}
    \label{tab:ablation_study}
    \resizebox{\columnwidth}{!}{%
    \begin{tabular}{lccc}
    \toprule
    \textbf{Configuration} & \textbf{ES} & \textbf{EM} & \textbf{Ratio} \\
    \midrule
    \cellcolor{cyan!10}\approach & \cellcolor{cyan!10}\textbf{57.55} & \cellcolor{cyan!10}\textbf{32.40} & \cellcolor{cyan!10}4.3x \\
    \midrule
    \textbf{Coarse-grained Ablations:} & & & \\
    \quad w/ Similarity-based Ranking & 49.66 \textcolor{red}{\scriptsize(-7.89)} & 25.20 \textcolor{red}{\scriptsize(-7.20)} & 4.3x \\
    \quad w/ Random Ranking & 39.76 \textcolor{red}{\scriptsize(-17.79)} & 11.50 \textcolor{red}{\scriptsize(-20.90)} & 4.4x \\
    \midrule
    \textbf{Fine-grained Ablations:} & & & \\
    \quad w/o Fine-grained Compression & 56.10 \textcolor{red}{\scriptsize(-1.45)} & 31.20 \textcolor{red}{\scriptsize(-1.20)} & 4.2x \\
    \quad w/o Adaptive Budget Allocation & 55.21 \textcolor{red}{\scriptsize(-2.34)} & 29.40 \textcolor{red}{\scriptsize(-3.00)} & 4.3x \\
    \quad w/ Line Chunking & 55.98 \textcolor{red}{\scriptsize(-1.57)} & 31.20 \textcolor{red}{\scriptsize(-1.20)} & 4.3x \\
    \quad w/ Random Line Selection & 55.07 \textcolor{red}{\scriptsize(-2.48)} & 29.00 \textcolor{red}{\scriptsize(-3.40)} & 4.3x \\
    \bottomrule
    \end{tabular}%
    }
    \vspace{-0.5cm}
\end{table}

\subsection{RQ2: Ablation Study}

To understand the contribution of each component in \approach, we conduct an ablation study on the Long Code Completion task using Qwen2.5-Coder-7B. For all ablations, the total token budget and other hyper-parameters are set the same as the full method. We systematically remove or modify key components to analyze their individual impact. For coarse-grained ablations, we replace our conditional perplexity-based ranking with similarity-based ranking, and compare against random function ranking to establish a lower bound. For fine-grained ablations, we test four variants: removing fine-grained compression entirely (coarse-grained selection only), removing adaptive budget allocation (uniform budget allocation), replacing meta-chunking with simple line-based chunking, and using random line selection within selected functions.

Different components contribute varying degrees to performance. The coarse-grained ranking mechanism is most critical - conditional perplexity-based ranking outperforms similarity-based approaches by 7.89\% and random selection by 17.79\% in ES score. This demonstrates that semantic relevance through conditional perplexity is superior to lexical similarity. For fine-grained components, adaptive budget allocation improves ES by 2.34\%, enabling important functions to retain more detail. Perplexity-based chunking outperforms simple line chunking by 1.57\% in ES while being more computationally efficient, as line-by-line compression ranking would incur higher overhead compared to block-based analysis. Knapsack-based selection outperforms random line selection by 2.48\% in ES, confirming relevance-guided selection helps compression quality.

\begin{finding}{Finding 2}
    Coarse-grained conditional perplexity ranking has the most impact on the performance of LongCodeZip, while fine-grained optimizations further improve the compression information density.
\end{finding}

\subsection{RQ3: Transferability}

\begin{table}[t]
    \centering
    \caption{Cross-model Results}
    \label{tab:cross_model}
    \resizebox{\columnwidth}{!}{
    \begin{tabular}{lccc|c}
    \toprule
    \textbf{Compression Model} & \textbf{DS-6.7B} & \textbf{Seed-8B} & \textbf{Qwen-7B} & \textbf{Avg. ES} \\
    \midrule
    \textit{No Compression}    & 57.14 & 64.04 & 56.36 & 59.18 \\
    \textit{No Context}        & 41.29 & 41.88 & 38.14 & 40.44 \\
    \midrule
    \textsc{Deepseek-Coder-6.7B} & 60.58 & 61.48 & 56.55 & 59.54 \\
    \textsc{Seed-Coder-8B}       & 60.86 & \textbf{63.11} & 55.95 & 59.97 \\
    \textsc{Qwen2.5-Coder-0.5B}  & 61.12 & 62.68 & 56.58 & 60.13 \\
    \textsc{Qwen2.5-Coder-1.5B}  & 60.89 & 62.79 & 56.18 & 59.95 \\
    \textsc{Qwen2.5-Coder-3B}    & 60.74 & 63.10 & 56.79 & 60.21 \\
    \textsc{Qwen2.5-Coder-7B}    & \textbf{61.34} & 62.62 & \textbf{57.55} & \textbf{60.58} \\
    \bottomrule
    \end{tabular}%
    \vspace{-1cm}
    }
\end{table}

\begin{table*}[t]
    \begin{center}
    \caption{Efficiency Analysis of Different Methods}
    \label{tab:efficiency_analysis}
    \resizebox{0.9\textwidth}{!}{
    \begin{tabular}{lccccccc}
    \toprule
    \textbf{Method} & \textbf{Comp. Time (s)} & \textbf{Comp. GPU Mem (GB)} & \textbf{Gen. Time (s)} & \textbf{Gen. GPU Mem (GB)} & \textbf{Ratio} & \textbf{ES} & \textbf{EM} \\
    \midrule
    No Compression & 0.0 & 0.0 & 15.70 & \textit{Base} + 3.48 & 1.0x & 56.36 & 31.80 \\
    No Context & 0.0 & 0.0 & 0.68 & \textit{Base} + 0.06 & - & 38.14 & 9.60 \\
    \midrule
    RAG (Function Chunking) & 0.53 & 1.07 & 7.57 & \textit{Base} + 1.13 & 3.1x & 52.79 & 26.00 \\
    LLMLingua-2 & 0.65 & 4.71 & 6.53 & \textit{Base} + 0.79 & 4.4x & 41.29 & 12.20 \\
    DietCode & 15.23 & 0.0 & 7.26 & \textit{Base} + 1.03 & 3.4x & 43.91 & 13.20 \\
    SlimCode & 0.35 & 0.0 & 6.48 & \textit{Base} + 0.78 & 4.5x & 40.85 & 12.20 \\
    \midrule
    \cellcolor{cyan!10}\textbf{\approach} & \cellcolor{cyan!10}2.58 & \cellcolor{cyan!10}\textit{Base} + 0.69 & \cellcolor{cyan!10}6.59 & \cellcolor{cyan!10}\textit{Base} + 0.81 & \cellcolor{cyan!10}4.3x & \cellcolor{cyan!10}\textbf{57.55} & \cellcolor{cyan!10}\textbf{32.40} \\
    \bottomrule
    \end{tabular}
    }
    \begin{minipage}{0.88\textwidth}
    \vspace{0.5em}
    \footnotesize{\textit{Note}: Comp.: Compression, Gen.: Generation, Mem: Memory. \textit{Base} model parameters memory: 28.37 GB.}
    \end{minipage}
    \vspace{-0.5cm}
    \end{center}
\end{table*}

Table~\ref{tab:cross_model} presents the cross-model performance (ES) of our approach in the long code completion task.
Each row denotes the model used for context compression, while each column specifies the model used for code generation given the compressed context as the input. The results show that our approach generalizes well across different model architectures and sizes, regardless of which compression or generation model is used for downstream tasks. Models released at different times, from DeepSeek-Coder in 2023~\cite{guo2024deepseek} to Qwen2.5-Coder in 2024~\cite{hui2024qwen2} and Seed-Coder in 2025~\cite{zhang2025seed}, achieve similarly strong performance with only minor differences in average ES scores. Notably, even small models (e.g., Qwen2.5-Coder-0.5B) are highly effective, highlighting the strong transferability of our method. \edit{Using such small models will significantly reduce compression time and memory overhead, making our approach particularly suitable for resource-constrained scenarios.}

\begin{finding}{Finding 3}
    Our \approach generalizes well across different types and sizes of models in the cross-model setting, using a 0.5B model can also bring promising performance.
\end{finding}

\subsection{RQ4: Efficiency Analysis}
To evaluate the practical efficiency of \approach, we analyze the Long Code Completion task using Qwen2.5-Coder-7B by measuring both compression overhead and downstream benefits. We select several representative baselines based on their downstream performance in Table~\ref{tab:efficiency_analysis}. 
The GPU memory costs represent peak memory usage per stage, with generation memory cost referring to additional memory for forward propagation during generation beyond base model parameters (28.37GB). Due to the space limit, we only report the results with several representative baselines. Note that SlimCode and DietCode require no GPU memory for compression because they are not based on neural models.

Table~\ref{tab:efficiency_analysis} demonstrates that \approach achieves superior compression efficiency while maintaining the best performance. While our method requires a slightly higher compression overhead of 2.58s and additional GPU memory compared to the baselines, it significantly reduces input token costs by 77\% and decreases generation latency from 15.70s to 6.59s compared to no compression. This also translates to substantial cost savings when using expensive commercial LLM APIs, where pricing is primarily based on input token count. More importantly, as demonstrated in RQ3, the compression overhead can be effectively mitigated by using a lightweight 0.5B model without sacrificing quality. And the efficiency gains can also be further enhanced through techniques like quantization~\cite{frantar2023gptq}, making our approach highly practical for real-world deployment scenarios where cost efficiency is paramount.

\begin{finding}{Finding 4}
    Our \approach achieves 4.3× compression ratio with only 2.6s overhead, reduces generation time from 15.7s to 6.6s, yet it still maintains high downstream performance. 
\end{finding}

\begin{figure}[t]
    \centering
    \includegraphics[width=0.7\columnwidth]{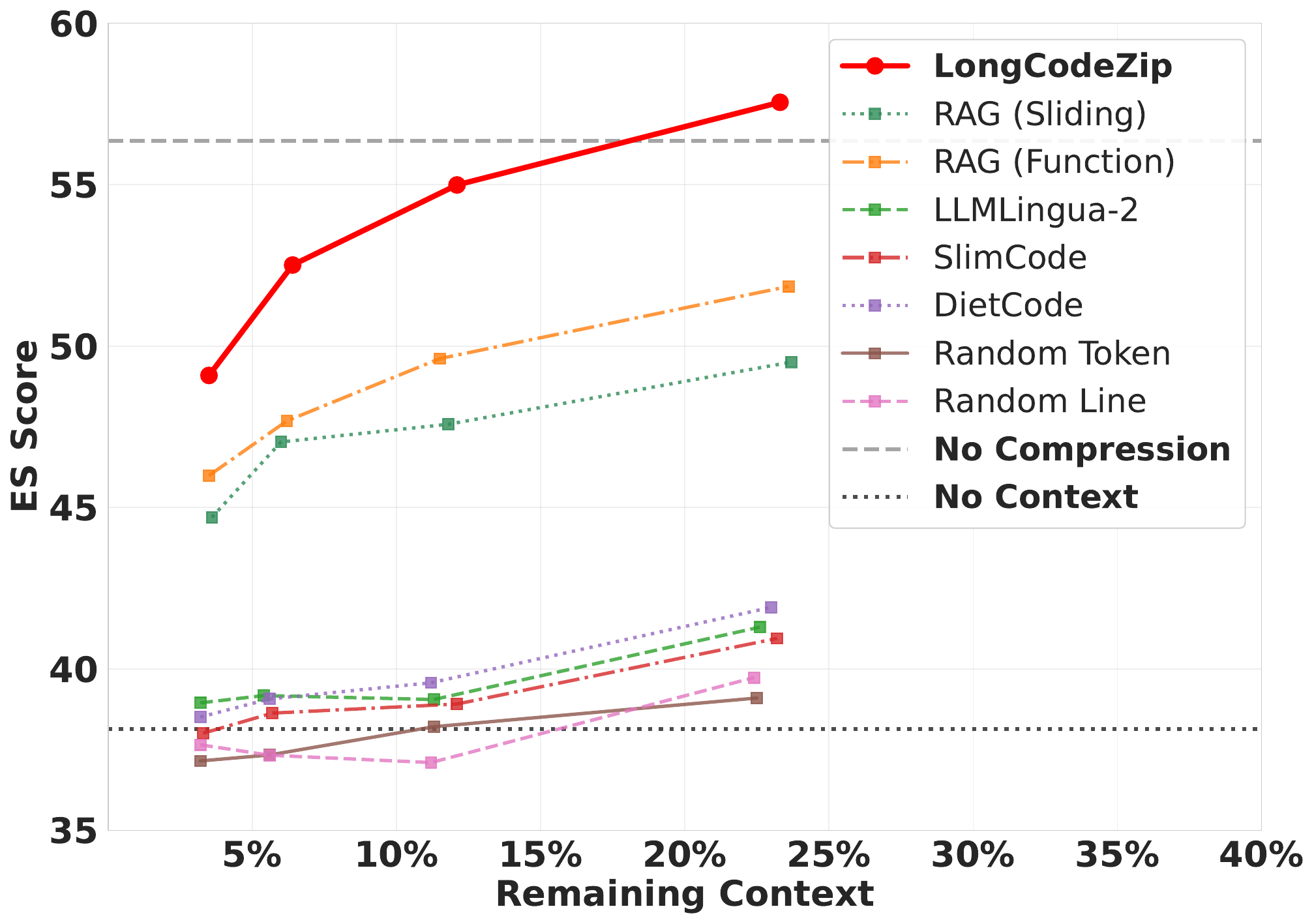}
    \caption{Performance (ES) vs remaining context (\%).} 
    \label{fig:compression_ratio}
    \vspace{-0.5cm}
\end{figure}

\begin{figure*}[t]
    \centering
    \includegraphics[width=0.9\textwidth]{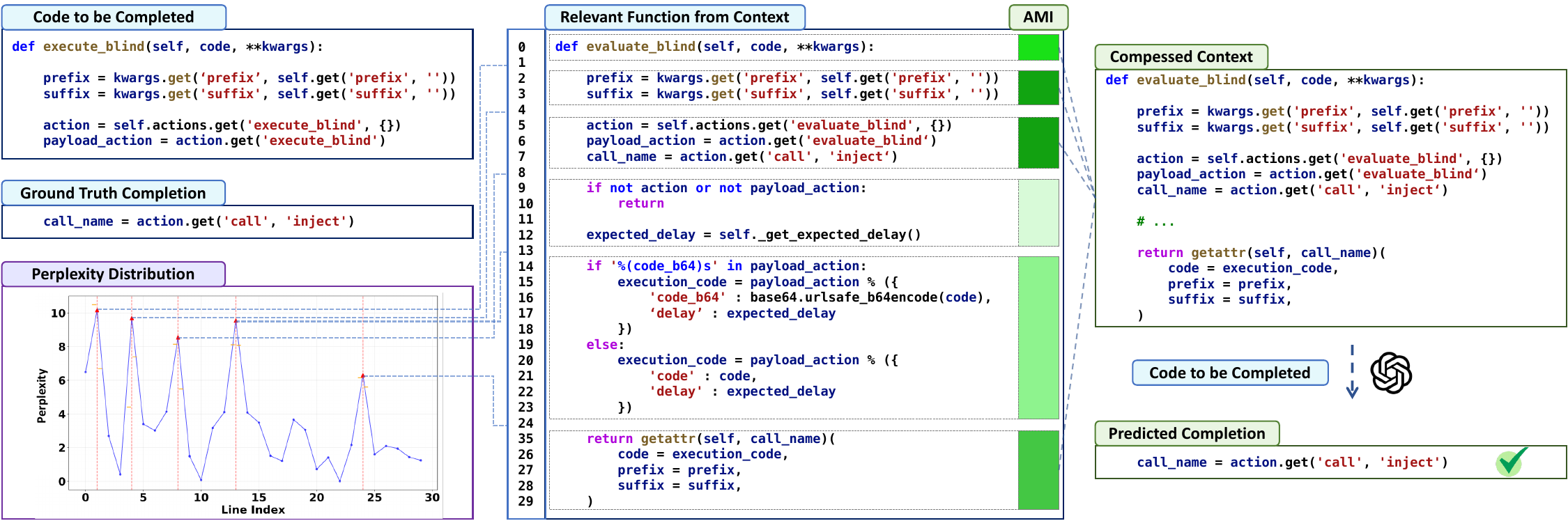}
    \caption{Example of fine-grained compression process on long code completion.}
    \label{fig:case}
    \vspace{-0.5cm}
\end{figure*}


\section{Discussion}


\subsection{Compression vs Performance}
Understanding the relationship between compression ratio and model performance is essential for evaluating the effectiveness of code compression methods in long-context scenarios.
Figure~\ref{fig:compression_ratio} presents the ES score versus the percentage of remaining context, showing Qwen2.5-Code-7B results for representative methods on the Long Code Completion task. LongCodeZip consistently achieves the highest ES scores across all compression ratios, demonstrating its strong ability to identify and retain the most relevant context for code completion. Notably, LongCodeZip can effectively leverage additional context, resulting in substantial performance gains—especially at severe compression ratios (with remaining context less than 10\%) where context is extremely limited. This gain becomes less pronounced at more relaxed compression ratios, which is reasonable since our method ranks and selects the most relevant functions early on, so the marginal benefit of extra context diminishes. In contrast, most baselines perform close to random selection, and adding more context does not significantly improve their ES scores. Among the baselines, RAG-based methods do exhibit improvement as more context is retained, but their overall ES scores remain significantly lower than those of LongCodeZip.


\subsection{Case Study}

We illustrate the effectiveness of \approach through a case study in Figure~\ref{fig:case}, focusing on the fine-grained compression stage (coarse-grained design choice is discussed in Section~\ref{sec:motivation}). Our method identifies semantic boundaries at positions where a line's perplexity sharply increases relative to its neighbors. The method tends to separate out major, independent functional modules, and also naturally groups together smaller, closely related segments. Some of our detected boundaries align with double newlines, which is consistent with common programming practices in codes with good code smell. 
The resulting compressed blocks, highlighted by boxes in the right panel, preserve the key information need for completion. The preserved blocks closely matches the code to be completed. This shows our approach effectively compresses code while retaining relevant and useful content to the task.

In our experiments, we have also observed some common failure modes.
In particular, when the context either lacks information relevant to the task instruction or when it is difficult to align an ambiguous instruction with any segment of the context, our method may struggle to identify and preserve useful blocks. 

\subsection{\edit{Necessity of Two-staged Compression}}
\edit{While the coarse-grained step provides the largest compression gains by removing entire irrelevant functions, the fine-grained compression step is crucial for balancing the trade-off between compression overhead and task model cost. Users can disable the fine-grained step for faster, cheaper compression when using less expensive models. However, for powerful but costly APIs like Claude-3.7-Sonnet, the precise pruning from the fine-grained step becomes critical, yielding substantial cost reductions that justify the additional computational overhead. This adaptive design allows \approach to accommodate different deployment scenarios and cost constraints.}

\section{Threats to Validity}


While our evaluation is comprehensive, several threats to validity should be acknowledged. 1) A primary threat concerns the accuracy of our evaluation for the summarization evaluation relies on LLM-generated scores, which may differ from human expert assessments and potentially suffer from ordering effects. To mitigate this, we followed the original paper~\cite{bogomolov2024long} to average scores over different prompt orderings and employed GPT-4o-mini as an independent referee. These practices reduce bias and improve the objectivity of our results.
2) Our findings may be specific to the datasets, programming languages, or LLMs used. To improve generalizability, we evaluated our approach across diverse datasets, languages, model families, and in cross-model settings. This diversity provides convincing evidence on the generalizability of our findings.
3) There is a risk of data leakage if models are exposed to benchmark data during training. To exclude potential effects of data leakage,  we used DeepSeek-Coder-6.7B~\cite{guo2024deepseek}, which was trained only on data available before March 2023, while all the benchmarks we evaluated were released after that date~\cite{guo2023longcoder, bogomolov2024long, liu2024repoqa}. This step helps ensure the integrity and reliability of our results.

\section{Related Work}
\subsection{Large Language Models for Code}
General-purpose LLMs such as GPT~\cite{achiam2023gpt}, Gemini~\cite{team2023gemini, team2024gemini}, Qwen~\cite{bai2023qwen, yang2024qwen2} and DeepSeek~\cite{guo2025deepseek} demonstrate strong code capabilities through large-scale pretraining on diverse data. To better perform on code-related tasks, a series of code-specialized LLMs have been proposed. CodeX~\cite{chen2021evaluating} adapts GPT architecture and is pretrained on a large corpus of GitHub code using next-token prediction. Similarly, CodeGen~\cite{nijkamp2022codegen} adopts a decoder-only architecture, with a focus on multi-turn program synthesis and open-source availability. StarCoder~\cite{li2023starcoder}, on the contrary, adopts a fill-in-the-middle objective for improved bidirectional context modeling. CodeLlama~\cite{roziere2023code} extends LLaMA with code-specific tokenization and longer contexts. CodeT5+\cite{wang2023codet5+} employs span denoising in an encoder-decoder framework. More recent models further incorporate instruction tuning and reinforcement learning (RL) to improve alignment and generalization. WizardCoder~\cite{luo2023wizardcoder} fine-tunes StarCoder~\cite{li2023starcoder} with Evol-Instruct and ChatGPT~\cite{openai2022chatgpt} feedback. DeepSeek-Coder~\cite{guo2024deepseek} combines instruction tuning, RL and compiler feedback to optimize for correctness and human preference. Qwen2.5-Coder~\cite{hui2024qwen2} also undergoes instruction tuning and RL, with a focus on long-context fidelity through multi-stage alignment.

These models have demonstrated remarkable performance in downstream tasks such as code generation~\cite{nijkamp2022codegen,shi2024between}, code summarization~\cite{ahmad2020transformer,wang2025context}, and code question answering~\cite{gu2018deep,peng2025swe}.
Despite increasingly longer context windows, LLMs exhibit significant limitations in long context scenarios, especially when relevant information appears in the middle of a prompt~\cite{liu2023lost, li2023loogle} or when code completion requires cross-function or structural dependencies~\cite{bogomolov2024long, yu2024codereval}.
To address these challenges, a number of long code benchmarks have been introduced, such as LongCodeBench~\cite{rando2025longcodebench}, LongCodeU~\cite{li2025longcodeu}, YABLoCo~\cite{valeev2025yabloco}, and LongCodeArena~\cite{bogomolov2024long}. 
In parallel, recent research has tailored models to code tasks specifically. LongCoder~\cite{guo2023longcoder} adopts a sliding window mechanism for self-attention to enhance long-context code completion. HiRoPE~\cite{zhang2024hirope} leverages the hierarchical structure of source code to enable length extrapolation without additional training. aiXcoder-7B-v2~\cite{li2025aixcoder} introduces reinforcement learning-based fine-tuning to guide LLMs in utilizing long-range context for repository-level code completion. Complementing these architectural and training advancements, we propose an efficient code context compression technique that preserves essential semantic information while enabling LLMs to operate effectively within constrained input lengths.

\subsection{Context Compression of Large Language Models}

Context compression strategies can be categorized into hard prompt methods and soft prompt methods~\cite{li2024prompt}.
Soft prompt methods~\cite{mu2023learning, li2024500xcompressor, wang2024adapting} summarize input into dense vectors or prefix embeddings. While memory-efficient, they require fine-tuning on the target model, making them impractical in closed-source settings like using GPT-4. In contrast, hard prompt methods directly manipulate the input by removing or rephrasing less informative content. LLMLingua~\cite{jiang2023llmlingua}, LongLLMLingua~\cite{jiang2024longllmlingua}, and Selective Context~\cite{li2023compressing} use learned or statistical importance scores to prune uninformative tokens or sentences. LLMLingua-2~\cite{pan2024llmlingua} advances this by employing data distillation from GPT-4 to train a token classification model, achieving efficient and faithful compression. AttentionRAG~\cite{fang2025attentionrag} prunes the context based on the attention between queries and retrieved contexts.

However, these methods are primarily designed for natural language, and often fail to capture the structural and semantic regularities of source code, leading to suboptimal performance in code-related tasks. This has led to a growing body of research focused on code-specific compression. ShortenDoc~\cite{yang2024less} targets docstring compression specifically, whereas our method targets source code, which typically dominates the input in long-context scenarios. DietCode~\cite{zhang2022diet} combines static frequency-based filtering with CodeBERT attention heuristics to discard low-impact tokens, but its reliance on model-specific attention reduces adaptability across different architectures. SlimCode\cite{wang2024natural} applies rule-based token pruning using token types and program dependency graphs, which may not generalize well across languages or tasks. 
\edit{However, these existing methods mainly focus on compressing single functions for short context tasks.

Additionally, advanced RAG-based approaches have been developed for enhancing repository-level code completion by retrieving relevant context, including A\textsuperscript{3}-CodGen~\cite{liao2024a3codgen}, which incorporates third-party library information; cAST~\cite{zhang2025cast}, which leverages structural chunking via abstract syntax trees; RepoGenix~\cite{liang2024repogenix}, which combines analogous and relevant contexts; and RLCoder~\cite{wang2024rlcoder}, which trains stronger retrievers for improved context selection. Unlike these approaches that are specifically designed for repository-level code completion, we propose a training-free code context compression technique that provides broader applicability across diverse long-context code tasks including summarization and question answering by preserving essential semantic information while enabling existing LLMs to operate effectively within constrained input lengths. Our contribution lies in the synergistic integration and significant code-aware approach to address the unique structural and semantic characteristics of programming languages. 

To the best of our knowledge, our approach is the first to explicitly target long-context compression in code LLMs, providing a training-free and model-agnostic solution that efficiently preserves task-relevant content under tight token budgets.}

\section{Conclusion}

In this paper, we have introduced \approach, a training-free, model-agnostic and plug-and-play framework for long-context code compression. Our two-stage hierarchical approach combines function-level selection with block-level pruning.
Comprehensive experiments across code completion, summarization, and question answering tasks demonstrate that \approach achieves up to a 5.6x compression ratio without sacrificing task performance, consistently outperforms existing baselines, and significantly reduces computational costs. The framework exhibits strong cross-model generalization and maintains competitive performance even with a lightweight 0.5B compression model. As the first framework specifically designed for long-context code compression, \approach enables code LLMs to scale more efficiently to real-world, large-scale software development scenarios.


\section*{Acknowledgment}
We thank anonymous reviewers for their constructive comments. We also thank Maoquan Wang and Yanfu Yan for helpful discussions on improving this work.
This research is funded by the National Key Research and Development Program of China (Grant No. 2023YFB4503802) and the Natural Science Foundation of Shanghai (Grant No. 25ZR1401175).






\bibliographystyle{IEEEtran}
\bibliography{reference}
\balance
\end{document}